\documentclass{article}
\usepackage{spconf,amsmath, amssymb,graphicx,hyperref,bm,multirow,booktabs,arydshln,courier,caption,nicematrix, tikz,dblfloatfix, enumitem, url}

\title{A Decomposition-based State Space Model for \\ Multivariate Time-Series Forecasting}
\name{Shunya Nagashima, Shuntaro Suzuki, Shuitsu Koyama, Shinnosuke Hirano\thanks{Our code is available at \href{https://github.com/Neurogica/DecompSSM}{https://github.com/Neurogica/DecompSSM}}}

\address{Neurogica Inc.}
\begin{document}
%
\maketitle
\begin{abstract}

Multivariate time series (MTS) forecasting is crucial for decision-making in domains such as weather, energy, and finance. It remains challenging because real-world sequences intertwine slow trends, multi-rate seasonalities, and irregular residuals. Existing methods often rely on rigid, hand-crafted decompositions or generic end-to-end architectures that entangle components and underuse structure shared across variables. To address these limitations, we propose DecompSSM, an end-to-end decomposition framework using three parallel deep state space model branches to capture trend, seasonal, and residual components. The model features adaptive temporal scales via an input-dependent predictor, a refinement module for shared cross-variable context, and an auxiliary loss that enforces reconstruction and orthogonality. Across standard benchmarks (ECL, Weather, ETTm2, and PEMS04), DecompSSM outperformed strong baselines, indicating the effectiveness of combining component-wise deep state space models and global context refinement.
\end{abstract}
\newcommand{\parhead}[1]{\vspace{0.35\baselineskip}\noindent\textbf{#1}\hspace{0.5em}}

\vspace{-2mm}
\section{Introduction}
\label{sec:introductoin}
\vspace{-1mm}
Multivariate time series (MTS) forecasting is crucial for decision-making in a wide range of fields, including weather, energy, and finance~\cite{Wu_intro_weather_2023, Martin_intro_energy_20101772}. 
A fundamental challenge is that real-world series are often a composite of multiple underlying patterns, such as long-term trends and various periodic fluctuations occurring simultaneously. 

Accordingly, a central principle in time series analysis has been the decomposition of a series into these constituent components~\cite{anderson_tsf_2011, cleveland_tsf_90}.
However, attempts to integrate this principle into modern architectures have primarily followed three directions, each with inherent limitations: 
(1) A predominant strategy, exemplified by DLinear~\cite{zeng_dlinear_2023}, MICN~\cite{wang_micn_2023}, Autoformer~\cite{wu_autoformer_2021}, FEDformer~\cite{zhou_fedformer_2022}, and TimeMixer~\cite{wang_timemixer_2024}, is to rely on a predefined, rigid moving average, which is non-adaptive and risks discarding information. 
(2) A second group (e.g., LaST~\cite{wang_last_2022}, CoST~\cite{woo_cost_2022}) learns decomposition in latent space, but uses generic, non-specialized encoder architectures.
(3) A final category (e.g., PPDformer~\cite{wan_ppdformer_2025}) uses a separate pre-processing pipeline to identify and isolate periodic components. However, this form of decomposition is not end-to-end trainable and thus can not adapt to the forecasting objective.

\noindent
\begin{list}{$\bullet$}{\setlength{\leftmargin}{1em}\setlength{\itemsep}{0.25ex}\setlength{\topsep}{0.25ex}}
\raggedright
\item We propose DecompSSM that decomposes trend, seasonal, and residual structure using parallel branches with input-dependent, component-wise timescale adaptation.
\item We introduce a Global Context Refinement module that feeds back shared cross-variable context, re-synchronizing variables and reducing misassignment under noise/missingness.
\item We design an Auxiliary Decomposition Loss to promote faithful component decomposition by enforcing additive reconstruction and pairwise orthogonality.
\item Across ECL, Weather, ETTm2, and PEMS04 datasets, our method outperformed all baselines in most settings.
\end{list}

\vspace{-3mm}
\section{Related Work}
\label{sec:related_work}
\vspace{-2mm}

Recent advances in deep learning for time series forecasting are comprehensively detailed in \cite{benidis_tsf_survey_2022, wang_tsf_survey_2024}.
Furthermore, deep state space models (SSMs), a core component of our approach, have emerged as a powerful and efficient alternative to Transformer-based architectures, with their recent progress comprehensively summarized in \cite{patro_ssm_survey_2025, wang_ssm_survey_2024}.

\begin{figure*}[t]
  \centering
  \includegraphics[width=0.87\linewidth]{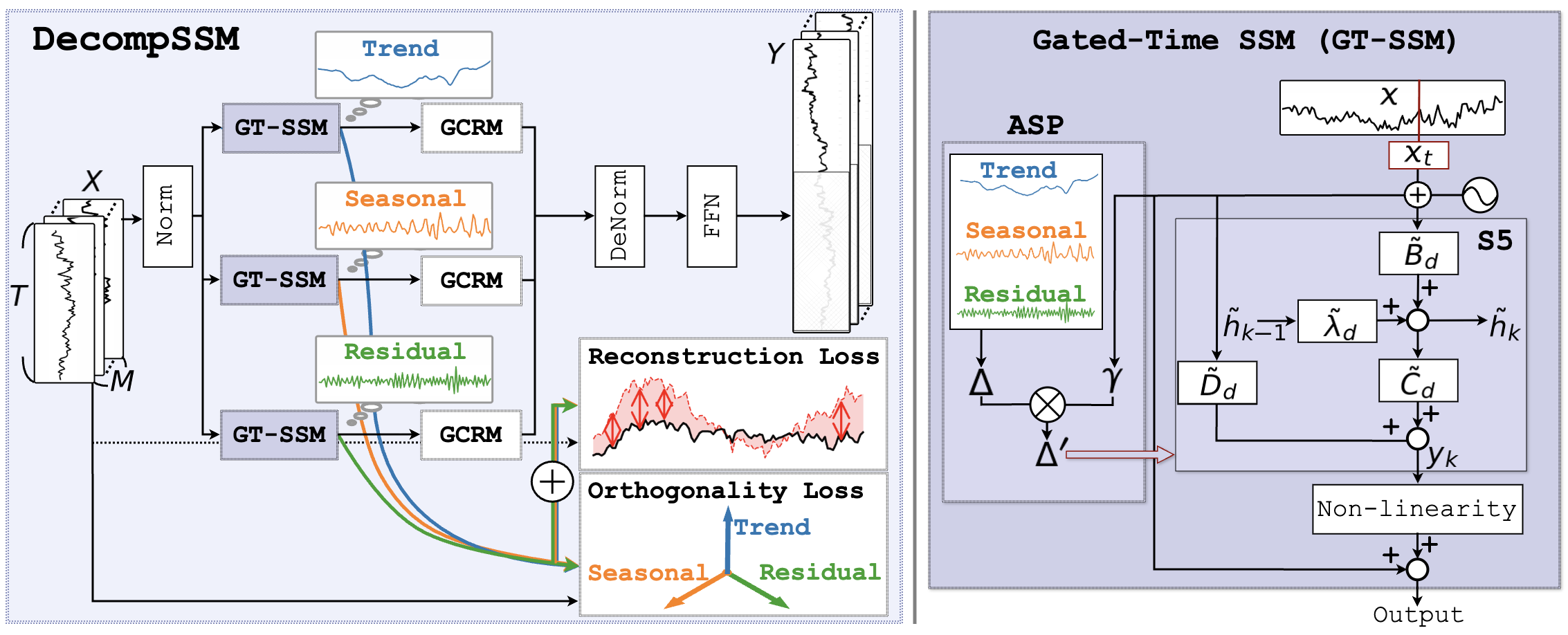}
  \vspace{-2mm}
  \caption{Overview of DecompSSM. Left: decomposition-based forecasting with three GT-SSM branches (trend, seasonal, residual) and auxiliary objectives (orthogonality, reconstruction). Right: the Gated-Time SSM (GT-SSM) with the input-dependent Adaptive Step Predictor (ASP) built on S5\cite{smith_s5_2023}.}
  \vspace{-2mm}
  \label{fig:model}
\end{figure*}

\parhead{Multivariate time series forecasting.}
Modern deep learning methods for time series forecasting increasingly attempt to incorporate the foundational principle of decomposition.
A primary strategy is to apply a predefined decomposition to the input series, an approach utilized by Transformer-based methods such as Autoformer and FEDformer. This same simplistic moving-average decomposition is also adopted by models with different core architectures, such as the MLP-based TimeMixer. 
Another approach focuses on learning disentangled representations; for instance, LaST employs variational inference to infer distinct seasonal and trend components. 
In contrast, methods such as PPDformer use a fixed, non-end-to-end pre-processing pipeline to isolate periodic features based on short-time Fourier transform.
However, these diverse approaches highlight that a key challenge remains in developing a decomposition framework that is simultaneously adaptive, architecturally specialized, and end-to-end trainable.

\parhead{Deep state space models.}
SSMs have emerged as an efficient alternative to Transformers for long-sequence modeling. 
Recent advances include Mamba~\cite{Gu_Mamba_2024}, which introduces selective state spaces with input-dependent parameterization, enabling linear-time scaling and high throughput across modalities; its success has spurred adoption in computer vision and speech~\cite{Zhu_vision_mamba_2024,wang_mambareg_2024,jiang_dpmamba_2025}.
Mamba-2~\cite{dao_mamba2_2024} formalizes the SSM–masked-attention duality and reports \(2\text{–}8\times\) speedups over the original selective scan while supporting larger state sizes and longer contexts. 
However, while Mamba and Mamba-2 leverage input-conditioned selection and efficient scans, we adopt S5~\cite{smith_s5_2023} because its multi-input multi-output (MIMO) parameterization explicitly models cross-variable dependencies within the state transition, which is particularly suitable for MTS forecasting.

\vspace{-3mm}
\section{Proposed Method}
\label{sec:methods}
\vspace{-2mm}
\begin{table*}[!t] 
\centering
\caption{MTS forecasting results, where the best results are in \textbf{bold}, the second best results are \underline{underlined}.}
\vspace{-3mm}
\label{tab:quantitative_results}
\resizebox{0.92\textwidth}{!}{%
\begin{tabular}{l|c|c@{\hspace{6pt}}c|c@{\hspace{6pt}}c|c@{\hspace{6pt}}c|c@{\hspace{6pt}}c|c@{\hspace{6pt}}c|c@{\hspace{6pt}}c|c@{\hspace{6pt}}c|c@{\hspace{6pt}}c}
\toprule
\multicolumn{2}{l|}{\textbf{Method}} & \multicolumn{2}{c|}{\textbf{\begin{tabular}{@{}c@{}}Ours \\ (2025)\end{tabular}}} & \multicolumn{2}{c|}{\textbf{\begin{tabular}{@{}c@{}}PPDformer \\ (2025)\end{tabular}}} & \multicolumn{2}{c|}{\textbf{\begin{tabular}{@{}c@{}}HDMixer \\ (2024)\end{tabular}}} & \multicolumn{2}{c|}{\textbf{\begin{tabular}{@{}c@{}}iTransformer \\ (2024)\end{tabular}}} & \multicolumn{2}{c|}{\textbf{\begin{tabular}{@{}c@{}}PatchTST \\ (2023)\end{tabular}}} & \multicolumn{2}{c|}{\textbf{\begin{tabular}{@{}c@{}}TimesNet \\ (2023)\end{tabular}}} & \multicolumn{2}{c|}{\textbf{\begin{tabular}{@{}c@{}}DLinear \\ (2023)\end{tabular}}} & \multicolumn{2}{c}{\textbf{\begin{tabular}{@{}c@{}}Autoformer \\ (2022)\end{tabular}}} \\
\midrule
\multicolumn{2}{l|}{\textbf{Metric}} & \textbf{MSE} & \textbf{MAE} & \textbf{MSE} & \textbf{MAE} & \textbf{MSE} & \textbf{MAE} & \textbf{MSE} & \textbf{MAE} & \textbf{MSE} & \textbf{MAE} & \textbf{MSE} & \textbf{MAE} & \textbf{MSE} & \textbf{MAE} & \textbf{MSE} & \textbf{MAE} \\
\midrule
\multirow{4}{*}{\rotatebox{90}{\textbf{ECL}}} 
& 96  & \textbf{0.137} & \textbf{0.234} & \underline{0.140} & \underline{0.239} & 0.173 & 0.276 & 0.148 & 0.240 & 0.195 & 0.285 & 0.174 & 0.278 & 0.216 & 0.306 & 0.201 & 0.317 \\
& 192 & \textbf{0.157} & \textbf{0.251} & \underline{0.159} & \underline{0.257} & 0.185 & 0.286 & 0.168 & 0.259 & 0.199 & 0.289 & 0.199 & 0.302 & 0.215 & 0.309 & 0.222 & 0.334 \\
& 336 & \underline{0.174} & \textbf{0.267} & \textbf{0.172} & \underline{0.273} & 0.197 & 0.299 & 0.178 & 0.271 & 0.215 & 0.305 & 0.204 & 0.305 & 0.228 & 0.323 & 0.231 & 0.338 \\
& 720 & \textbf{0.199} & \textbf{0.292} & \underline{0.200} & \underline{0.298} & 0.220 & 0.317 & 0.210 & 0.300 & 0.256 & 0.337 & 0.286 & 0.361 & 0.263 & 0.354 & 0.254 & 0.361 \\
\midrule
\multirow{4}{*}{\rotatebox{90}{\textbf{Weather}}} 
& 96  & \textbf{0.155} & \textbf{0.201} & \underline{0.160}& \underline{0.208} & 0.167 & 0.217 & 0.178 & 0.219 & 0.177 & 0.218 & 0.173 & 0.223 & 0.197 & 0.254 & 0.266 & 0.336 \\
& 192 & \textbf{0.204} & \textbf{0.246} & \underline{0.210} & \underline{0.254} & 0.222 & 0.261 & 0.226 & 0.259 & 0.225 & 0.259 & 0.219 & 0.263 & 0.237 & 0.294 & 0.307 & 0.367 \\
& 336 & \textbf{0.264} & \textbf{0.291} & \underline{0.268} & 0.297 & 0.278 & 0.301 & 0.282 & 0.300 & 0.275 & \underline{0.296} & 0.281 & 0.303 & 0.283 & 0.331 & 0.359 & 0.395 \\
& 720 & \textbf{0.345} & \textbf{0.343} & \underline{0.347} & \underline{0.347} & 0.353 & 0.350 & 0.358 & 0.350 & 0.354 & 0.348 & 0.357 & 0.352 & 0.348 & 0.383 & 0.419 & 0.428 \\
\midrule
\multirow{4}{*}{\rotatebox{90}{\textbf{ETTm2}}} 
& 96  & \textbf{0.174} & \textbf{0.253} & \underline{0.179} & \underline{0.263} & 0.182 & 0.267 & 0.185 & 0.272 & 0.179 & 0.264 & 0.188 & 0.268 & 0.195 & 0.294 & 0.215 & 0.297 \\
& 192 & \textbf{0.242} & \textbf{0.298} & \underline{0.250} & \underline{0.309} & 0.248 & 0.307 & 0.253 & 0.313 & 0.251 & 0.312 & 0.260 & 0.310 & 0.285 & 0.361 & 0.275 & 0.333 \\
& 336 & \textbf{0.300} & \textbf{0.334} & \underline{0.308} & \underline{0.342} & 0.310 & 0.346 & 0.317 & 0.352 & 0.311 & 0.346 & 0.321 & 0.349 & 0.392 & 0.433 & 0.331 & 0.367 \\
& 720 & \underline{0.406} & \textbf{0.396} & \textbf{0.403} & \underline{0.401} & 0.406 & 0.400 & 0.413 & 0.407 & 0.415 & 0.409 & 0.424 & 0.408 & 0.541 & 0.515 & 0.424 & 0.419 \\
\midrule
\multirow{4}{*}{\rotatebox{90}{\textbf{PEMS04}}} 
& 12  & \textbf{0.070} & \textbf{0.173} & \underline{0.077} & \underline{0.183} & 0.092 & 0.211 & 0.080 & 0.188 & 0.105 & 0.224 & 0.087 & 0.195 & 0.148 & 0.272 & 0.424 & 0.491 \\
& 24  & \textbf{0.084} & \textbf{0.191} & \textbf{0.084} & \underline{0.192} & 0.124 & 0.247 & 0.099 & 0.212 & 0.153 & 0.275 & 0.103 & 0.215 & 0.224 & 0.340 & 0.459 & 0.509 \\
& 48  & \textbf{0.108} & \textbf{0.221} & \underline{0.114} & \underline{0.228} & 0.181 & 0.302 & 0.131 & 0.245 & 0.229 & 0.339 & 0.136 & 0.250 & 0.355 & 0.437 & 0.646 & 0.610 \\
& 96  & \underline{0.149} & \underline{0.264} & \textbf{0.140} & \textbf{0.250} & 0.255 & 0.366 & 0.171 & 0.282 & 0.291 & 0.389 & 0.190 & 0.303 & 0.452 & 0.504 & 0.912 & 0.748 \\
\midrule
\multicolumn{2}{l|}{\textbf{Count}} & \multicolumn{2}{c|}{28} & \multicolumn{2}{c|}{5} & \multicolumn{2}{c|}{0} & \multicolumn{2}{c|}{0} & \multicolumn{2}{c|}{0} & \multicolumn{2}{c|}{0} & \multicolumn{2}{c|}{0} & \multicolumn{2}{c}{0} \\
\bottomrule
\end{tabular}%
}
\end{table*}
In this study, we address the task of MTS forecasting. The input is a multivariate time series $\bm{X} \in \mathbb{R}^{T \times M}$, where $T$ is the number of time steps and $M$ is the number of variables. The output is a sequence of predicted values for the next $H$ time steps, denoted as $\hat{Y} \in \mathbb{R}^{H \times M}$.

\vspace{-2mm}
\subsection{Overview}
\vspace{-1mm}
We propose DecompSSM, a method that performs MTS forecasting through a principled decomposition framework designed to be simultaneously adaptive, architecturally specialized, and end-to-end trainable. As illustrated in Figure\ref{fig:model}, the input series is processed in parallel by three specialized Gated-Time SSMs (GT-SSMs), each tasked with extracting a trend, seasonal, or residual component. Each GT-SSM incorporates an Adaptive Step Predictor (ASP) that modulates its discretization scale based on the input, enabling component-wise timescale adaptation. The outputs are then refined by a Global Context Refinement Module (GCRM), which injects shared cross-variable context via a residual pathway. The entire framework is trained end-to-end with a primary MSE loss, guided by two auxiliary losses: a Reconstruction Loss to enforce the decomposition principle, and an Orthogonality Loss to encourage the distinctness of the learned components.

\vspace{-2mm}
\subsection{Input Embedding}
\vspace{-1mm}
We first apply instance normalization to $\bm{X}$ to mitigate non-stationarity. Each normalized variate is then embedded into a $D$-dimensional representation, denoted as $\bm{X}' \in \mathbb{R}^{M \times D}$.
This $\bm{X}'$ is fed to the subsequent GT-SSM branches.


\vspace{-2mm}
\subsection{Gated-Time State Space Model}
\vspace{-1mm}
We propose GT-SSM, a core module designed to decompose MTS.
We use three parallel branches.
Let \(\mathcal{C}=\{\mathrm{tr},\mathrm{se},\mathrm{re}\}\) denote the set of components, each representing trend, seasonal, and residual components, respectively.
For each component \(c\in\mathcal{C}\), the branch takes input \(\bm{X}_c\) derived from the shared representation \(\bm{X}'\) and produces \(\bm{H}_c\in\mathbb{R}^{M\times D}\).

\parhead{Branch-wise positional embedding.}
Updates along the variable dimension can blur variable identity and component-specific differences. Small shifts in amplitude or phase may be misassigned across branches. To mitigate this, we add a branch-specific learnable positional embedding to the shared representation $\bm{X}'\!\in\!\mathbb{R}^{M\times D}$ and form the branch inputs $\bm{X}_{\mathrm{tr}}, \bm{X}_{\mathrm{se}}, \bm{X}_{\mathrm{re}}\in\mathbb{R}^{M\times D}$.

\parhead{Variate-centric SSM.}
GT-SSM applies its core SSMs across the variate dimension, departing from conventional models~\cite{wang_micn_2023, wu_autoformer_2021} that operate on the temporal dimension. The effectiveness of this variate-centric architecture has been demonstrated by models such as iTransformer~\cite{liu_itransformer_2024}. 
We build on the S5 layer, a variant of SSMs. Its MIMO parameterization is well suited to modeling temporal structure in MTS.


\parhead{Continuous-to-discrete formulation.}
A linear SSM is defined in continuous time by the differential equations:
\begin{equation}
\dot{\bm{h}}(t) = \bm{A}\bm{h}(t) + \bm{B}\bm{x}(t), \quad 
\bm{y}(t) = \bm{C}\bm{h}(t) + \bm{D}\bm{x}(t),
\end{equation}
where $\bm{h}(t)$, $\bm{x}(t)$, $\bm{y}(t)$, $\bm{A}$, $\bm{B}$, $\bm{C}$, and $\bm{D}$ are the state, input, output, state matrix, input matrix, output matrix, and feedthrough matrix, respectively. 
To be applied to a discrete sequence, the continuous-time parameters are converted into their discrete-time counterparts ($\bar{\bm{A}}$, $\bar{\bm{B}}$, $\bar{\bm{C}}$, $\bar{\bm{D}}$) using the Zero-Order Hold (ZOH) method with a timescale parameter $\Delta$.


\parhead{Adaptive Step Predictor.}
A key novelty of our approach is the ASP, which makes the timescale branch dependent. For each component \(c\), ASP learns a scaling factor by mean pooling over the feature dimension of \(\bm{X}_{c}\) and passing it through a two-layer FFN:
\begin{equation}
\begin{aligned}
\Delta_{\mathrm{scale}}^{(c)} &= \mathrm{FFN}\!\left(\mathrm{AvgPool}(\bm{X}_{c})\right),\\
\Delta^{\prime (c)} &= \Delta_{\mathrm{scale}}^{(c)}\,\Delta ,
\end{aligned}
\label{eq:asp}
\end{equation}
 where \(\log \Delta \sim \mathcal{U}\!\big(\log \Delta_{\min}, \log \Delta_{\max}\big)\) and \(\Delta_{\min},\Delta_{\max}\) bound the base timescale \(\Delta\). The modulated timescale \(\Delta^{\prime (c)}\) is then used for ZOH discretization in branch \(c\).

\parhead{Component-specific gating and frequency priors.}
Each branch applies a small gate: a two-layer MLP with a branch-specific nonlinearity and a final sigmoid to modulate the S5 output.
We use $\mathrm{Tanh}$ for trend (smooth), $\mathrm{GELU}$ for seasonal (quasi-periodic), and $\mathrm{ReLU}$ for residual (sparse/high-frequency).
We also initialize $(\Delta_{\min},\Delta_{\max})$ per branch to emphasize the intended band: wide for trend, intermediate for seasonal, and narrow for residual.

\vspace{-2mm}
\subsection{Global Context Refinement Module}
\vspace{-1mm}
In MTS forecasting, treating variables in isolation obscures the structure shared across variables that is needed to separate trend, seasonal, and residual patterns; without a shared reference, each variable learns its own scale and timing, and noise or missing values amplify these inconsistencies, causing variables to drift out of sync and components to be misassigned. To prevent this, we introduce GCRM, which aggregates a global summary over the variable dimension and feeds it back to each variable as a residual correction.

We apply the same operator \(\mathcal{A}\) to each component \(c\) in parallel, obtaining \(\mathbf{H}'_{c}=\mathcal{A}(\mathbf{H}_{c})\). For any \(\mathbf{H}\in\mathbb{R}^{M\times D}\), let \(\mathbf{g}\in\mathbb{R}^{D}\) be the mean of \(\mathbf{H}\) along the variable dimension, set \(\mathbf{z}=\mathbf{W}_g\,\mathbf{g}\) with \(\mathbf{W}_g\in\mathbb{R}^{D\times D}\), and form \(\mathbf{G}\in\mathbb{R}^{M\times D}\) by repeating \(\mathbf{z}\) across variables. The operator is
\begin{equation}
\mathcal{A}(\mathbf{H}) \;=\; \mathrm{LayerNorm}\!\big(\mathbf{H} + \sigma(\alpha)\,\mathbf{G}\big)\in\mathbb{R}^{M\times D},
\label{eq:gcrm}
\end{equation}
where \(\alpha\) is a learnable scalar and \(\sigma\) is the sigmoid function. LayerNorm is applied along the feature dimension.

\vspace{-2mm}
\subsection{Auxiliary Decomposition Loss}
\vspace{-1mm}
To promote decomposition, we penalize deviation from additive reconstruction and encourage pairwise orthogonality among the refined components. Let the component set be $\mathcal{C}=\{\mathrm{tr},\mathrm{se},\mathrm{re}\}$ and define the reconstructed input as $\widehat{\mathbf{X}}=\sum_{c\in\mathcal{C}}\mathbf{H}'_{c}$. 
Let $\rho(\mathbf{A},\mathbf{B})=\langle \mathbf{A},\mathbf{B}\rangle_F/(\|\mathbf{A}\|_F\|\mathbf{B}\|_F)$ denote the Frobenius cosine similarity.
We use two auxiliary terms:
\vspace{-1mm}
\begin{align}
\mathcal{L}_{\mathrm{rec}} &= \frac{1}{MD}\,\big\|\widehat{\mathbf{X}}-\mathbf{X}'\big\|_{F}^{2},\\
\mathcal{L}_{\mathrm{orth}} &= \sum_{c<d,\;c,d\in\mathcal{C}}\big|\rho(\mathbf{H}'_{c},\mathbf{H}'_{d})\big|.
\end{align}
The auxiliary decomposition loss is the weighted sum of these two terms and is added to the training objective.

The refined components are concatenated and fed to an FFN that maps them to the forecast horizon; the resulting sequence is de-normalized using the same per-variable statistics as in the normalization step, yielding $\hat{\mathbf{Y}}$. In training, we employ the MSE loss as the primary loss, and include an Auxiliary Decomposition Loss as a secondary term.
\vspace{-3mm}
\section{EXPERIMENTS}
\label{sec:experiments}
\vspace{-2mm}

\subsection{Setup}
\label{subsec:experimental_setup}
\parhead{Datasets.}
We evaluated our model on several standard datasets for MTS forecasting: ECL\footnote{\href{https://archive.ics.uci.edu/ml/datasets/ElectricityLoadDiagrams20112014}{https://archive.ics.uci.edu/ml/datasets/ElectricityLoadDiagrams20112014}}, Weather\footnote{\href{https://www.bgc-jena.mpg.de/wetter/}{https://www.bgc-jena.mpg.de/wetter/}}, ETT (ETTh1, ETTh2, ETTm1, and ETTm2)~\cite{Zhou_informer_2021} and PEMS (PEMS03, PEMS04, PEMS07, and PEMS08)~\cite{Chen_pems_2001}. 
Due to space constraints, this study primarily reports the results on the ETTm2 and PEMS04.

\parhead{Baselines.}
Seven baseline models were chosen for the evaluation of MTS forecasting, encompassing four Transformer-based models: Autoformer~\cite{wu_autoformer_2021}, PatchTST~\cite{nie_patchtst_2023}, iTransformer~\cite{liu_itransformer_2024}, and PPDformer~\cite{wan_ppdformer_2025}; two Linear-based models: DLinear~\cite{zeng_dlinear_2023} and HDMixer~\cite{huang_hdmixer_2024}; and a TCN-based model: TimesNet~\cite{wu_timesnet_2023}.

\parhead{Implementation details.}
We follow the experimental setup in prior work \cite{wan_ppdformer_2025} for fair comparison.
The prediction range is $H \in \{96,192,336,720\}$ and the look-back window $T$ is fixed at 96.
All experiments run on a single AWS g6e.xlarge instance equipped with one NVIDIA L40S GPU (48\,GB VRAM).
The number of training epochs is fixed to 10.
Each component branch uses a single-layer GT-SSM with a bidirectional S5 backbone.
We optimize the model with Adam.


\begin{table}[t]
\centering
\caption{Ablation on core modules of our method on the \textbf{Weather} dataset across four horizons.  Best is in \textbf{bold}; second best is \underline{underlined}.}
\vspace{-3mm}
\label{tab:ablation_module}
\setlength{\tabcolsep}{3pt}
\renewcommand{\arraystretch}{1.05}

\resizebox{0.95\columnwidth}{!}{
\begin{tabular}{c|cc|cc|cc|cc} 
\toprule
\textbf{Method} &
\multicolumn{2}{c|}{\textbf{Model (1-i)}} &
\multicolumn{2}{c|}{\textbf{Model (1-ii)}} &
\multicolumn{2}{c|}{\textbf{Model (1-iii)}} &
\multicolumn{2}{c}{\textbf{Model (1-iv)}} \\
\cmidrule(lr){1-1}\cmidrule(lr){2-9}
\textbf{Metric} & \textbf{MSE} & \textbf{MAE} & \textbf{MSE} & \textbf{MAE} & \textbf{MSE} & \textbf{MAE} & \textbf{MSE} & \textbf{MAE} \\
\midrule
96  & \textbf{0.155} & \textbf{0.201} & 0.184 & 0.223 & \underline{0.156} & \underline{0.202} & 0.159 & 0.205 \\
192 & \textbf{0.204} & \textbf{0.246} & 0.230 & 0.262 & \textbf{0.204} & \textbf{0.246} & \underline{0.207} & \underline{0.251} \\
336 & \textbf{0.264} & \textbf{0.291} & 0.285 & 0.301 & \underline{0.269} & \underline{0.293} & 0.271 & 0.297 \\
720 & \textbf{0.345} & \textbf{0.343} & 0.360 & 0.350 & 0.355 & 0.350 & \underline{0.353} & \underline{0.348} \\
\bottomrule
\end{tabular}
}
\end{table}

\begin{table}[t]
\centering
\caption{Ablation of sequence-modeling architectures on the \textbf{Weather} dataset across four horizons. Best is in \textbf{bold}; second best is \underline{underlined}.}
\vspace{-3mm}
\label{tab:ablation_ssm}
\setlength{\tabcolsep}{3pt}
\renewcommand{\arraystretch}{1.05}

\resizebox{0.95\columnwidth}{!}{
\begin{tabular}{c|cc|cc|cc|cc} 
\toprule
\textbf{Method} &
\multicolumn{2}{c|}{\textbf{Model (2-i)}} &
\multicolumn{2}{c|}{\textbf{Model (2-ii)}} &
\multicolumn{2}{c|}{\textbf{Model (2-iii)}} &
\multicolumn{2}{c}{\textbf{Model (2-iv)}} \\
\cmidrule(lr){1-1}\cmidrule(lr){2-9}
\textbf{Metric} & \textbf{MSE} & \textbf{MAE} & \textbf{MSE} & \textbf{MAE} & \textbf{MSE} & \textbf{MAE} & \textbf{MSE} & \textbf{MAE} \\
\midrule
96  & \textbf{0.155} & \textbf{0.201} & 0.165 & 0.211 & 0.161 & 0.208 & \underline{0.159} & \underline{0.203} \\
192 & \textbf{0.204} & \textbf{0.246} & 0.212 & 0.253 & 0.209 & 0.251 & \underline{0.208} & \underline{0.250} \\
336 & \textbf{0.264} & \textbf{0.291} & 0.271 & 0.296 & \underline{0.267} & \underline{0.293} & 0.268 & 0.296 \\
720 & \textbf{0.345} & \textbf{0.343} & \underline{0.353} & \underline{0.349}& 0.355 & 0.351 & 0.359 & 0.356 \\
\bottomrule
\end{tabular}
}
\end{table}

\vspace{-2mm}
\subsection{Quantitative Results}
\label{subsec:quantitative_results}

Table~\ref{tab:quantitative_results} shows the forecasting results. Rows labeled \{96, 192, 336, 720\} denote forecast horizons (steps ahead), and Count shows, for each method, the number of best entries across all dataset–horizon pairs for both MSE and MAE. 
Across the 32 settings, our method achieved the best score in 28 of them.
We used MSE and MAE as evaluation metrics, which are standard in MTS forecasting.
Averaged over all horizons and compared with the second best (PPDformer), our method outperformed it on every dataset:
ECL (MSE 0.6\%, MAE 2.2\%), Weather (MSE 1.7\%, MAE 2.3\%), ETTm2 (MSE 1.6\%, MAE 2.6\%), and PEMS04 (MSE 1.0\%, MAE 0.5\%).
The margins were most pronounced on ECL, Weather, and ETTm2, while PEMS04 also improved modestly.
These results indicate that explicitly separating trend, seasonal, and residual components and propagating global context across variables is effective across diverse datasets.

\vspace{-2mm}
\subsection{Ablation Study}
\label{subsec:quantitative_results}
\vspace{-2mm}

\parhead{Module-wise ablation.}
Table~\ref{tab:ablation_module} presents an ablation study on the core modules of DecompSSM using the Weather dataset. 
We compare four variants: (1-i) the proposed method, (1-ii) without GT-SSM branches, (1-iii) without ADL, and (1-iv) without GCRM. 
All variants show degraded performance compared to (1-i), with the largest drop observed when removing GT-SSM. 
These results indicate that the GT-SSM, which performs component-wise timescale adaptation, is the most critical module and that ADL and GCRM also provide meaningful contributions.


\parhead{Architecture ablation.}
Table~\ref{tab:ablation_ssm} presents an ablation on the sequence-modeling architectures using the Weather dataset. 
We compare four configurations: (2-i) our S5-based design, (2-ii) replacing S5 with Attention, (2-iii) replacing S5 with Mamba, and (2-iv) replacing S5 with Mamba-2. 
All alternatives show degraded performance compared to (2-i), with the largest drop observed when replacing S5 with Attention. 
These results indicate that S5, with its MIMO parameterization, is particularly well suited to MTS forecasting.

\vspace{-3mm}
\section{Conclusion}
\vspace{-2mm}

We propose DecompSSM, an MTS forecasting model that decomposes the series into trend, seasonal, and residual components using parallel branches featuring component-wise adaptive timescales, global context refinement, and a decomposition-guiding auxiliary loss.
Future work will focus on extending the framework to automatically determine the number of the branches from frequency bands, enabling its application to general signal-processing domains.

\bibliographystyle{IEEEbib}
\bibliography{strings,refs}

\end{document}